\documentclass{article}
\usepackage{colm2024_conference}

\usepackage{microtype}
\usepackage{hyperref}
\usepackage{url}
\usepackage{graphicx}
\usepackage{tabularx}
\usepackage{booktabs}
\usepackage{caption}
\usepackage{makecell}
\usepackage{booktabs}
\usepackage{multirow}
\usepackage{multicol}
\usepackage{colortbl}
\usepackage{amsmath}

\definecolor{darkblue}{rgb}{0, 0, 0.5}
\hypersetup{colorlinks=true, citecolor=darkblue, linkcolor=darkblue, urlcolor=darkblue}

\title{SDSAT: Accelerating LLM Inference through Speculative Decoding with Semantic Adaptive Tokens}

\author{Chengbo Liu\thanks{Thundersoft, \texttt{liucb0320@thundersoft.com}} \and Yong Zhu\thanks{Thundersoft, \texttt{zhuyong@thundersoft.com}, Corresponding Author}}

\begin{document}

\maketitle

\begin{abstract}
We propose an acceleration scheme for large language models (LLMs) through \underline{S}peculative \underline{D}ecoding with \underline{S}emantic \underline{A}daptive \underline{T}okens (SDSAT). The primary objective of this design is to enhance the LLM model's ability to generate draft tokens more accurately without compromising the model's accuracy. The core strategies involve: 1) Fine-tune the model by incorporating semantic adaptive tokens that possess flexible decoding capabilities without changing its structure, allowing them to generate high-quality draft tokens. 2) By employing a training method that does not affect the standard tokens, the model can acquire parallel decoding abilities atop its original framework with minimal training overhead. 3) We have designed the "two-step-draft-then-verify" generation strategies using both greedy search and nucleus sampling.
Experiments conducted on the CodeLlama-13B and 7B models have yielded speed increases of over 3.5X and 3.0X, respectively. Please refer to https://github.com/ainergy-ml/SDSAT
\end{abstract}

\section{Introduction}

Transformer-based LLMs, though efficiently trained in parallel on TPUs and GPUs, face limitations in auto-regressive sampling due to high memory bandwidth demands \citep{stern2018blockwise}, resulting in latency when generating multiple tokens one by one. Recently, algorithms based on the idea of speculative decoding have been developed specifically to optimize speed. These methods first obtain draft tokens at a lower cost, and then use the target model for verification. They can be categorized as follows:

The first category relies on smaller models. \cite{chen2023accelerating}, and \cite{leviathan2023fast} both utilize a smaller draft model that is faster but less powerful. The quality of the draft, generated by this model, is comparable to that of sampling a single token from the larger target model. \cite{leviathan2023fast} introduced the speculative sampling method and proved that the distribution of the generated text remains unchanged for both the greedy and non-greedy settings, achieving a 2X-3X acceleration compared to the standard T5X implementation.
\cite{zhou2023distillspec} leveraging Knowledge Distillation (KD) to enhance speculative decoding by aligning a smaller student (draft) model with a teacher (target) model for better acceptance rates.

Another category is to use the model itself, inferring a larger number of tokens at once to increase speed. \cite{santilli2023accelerating} implements the greedy decoding procedure in parallel with a block of tokens, without conducting any training. Medusa \citep{cai2023medusa} leverages a series of MLPs for token prediction, drawing on features from the second-to-top layer of the original Large Language Model. This method significantly reduces the time required to generate drafts. \cite{xia2023speculative} shows that their approach can achieve around a 5× speedup in various seq2seq tasks by masking tokens, including machine translation and abstractive summarization. However, the length of token generation processed is relatively short. Eagle \citep{li2024eagle} builds on the existing model by adding only a lightweight plug-in (a single transformer decoder layer) to the LLM. This plug-in is trained to bring the small model as close as possible to the capability of the target model. The target model's inference process uses tree attention \citep{cai2023medusa} \citep{miao2024specinfer} \citep{spector2023accelerating} to obtain all outputs at once, and the verification process uses speculative sampling \citep{leviathan2023fast} to achieve the final results from the target model.

The third category is data-driven draft token generation, represented by \cite{he2023rest}, which uses pre-made data to help quickly obtain alternative answers for new tasks, using the tree-attention method. This allows for the inference results of the target model to be obtained all at once.

Our proposed solution offers significant improvements without the need for introducing new smaller models or making modifications to the existing model. This eliminates the need for complex and difficult adaptation processes. Additionally, there is no requirement for an external database. By incurring a minimal additional training cost, the model can enhance its capability to generate highly accurate draft tokens, which in turn directly contributes to the acceleration of the model's performance.

\noindent The main contributions of this paper are as follows:

\textbullet{} We have verified that large language models (LLMs) can produce high-quality draft tokens without requiring any modifications to their structure, through the introduction of semantic adaptive tokens.

\textbullet{} We have developed an innovative training methodology that enables LLMs to produce accurate draft tokens without compromising the model's overall accuracy and performance.

\textbullet{} Furthermore, we propose an efficient "two-step-draft-then-verify" generation method for both greedy search and nucleus sampling \citep{holtzman2019curious}, which leads to high decoding efficiency.

\noindent Utilizing the CodeLlama-13B and 7B models \citep{roziere2023code}, and training only on 2-8B tokens, the model can maintain nearly unchanged accuracy while significantly enhancing speed. 

\begin{figure}[htbp]
    \centering
    \includegraphics[width=1\linewidth]{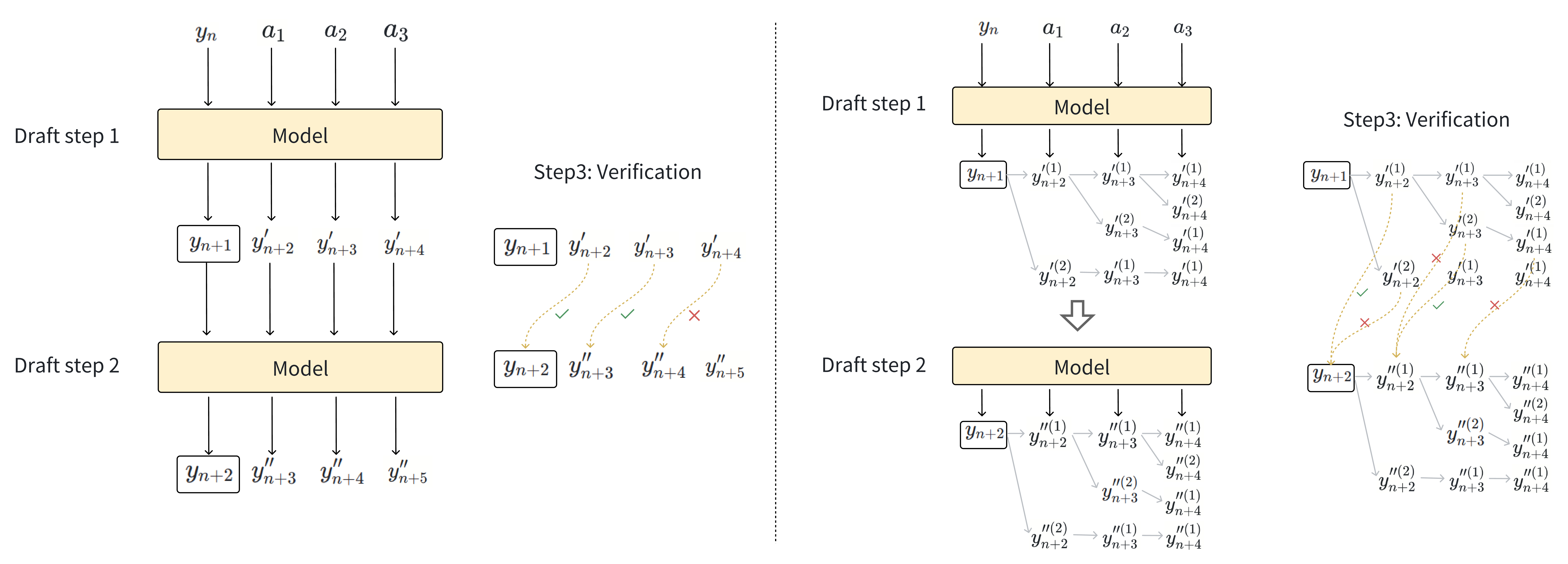}
    \caption{Diagram of the inference mechanisms, while the left figure depicts the greedy search process, which is a special pattern of the nucleus sampling process shown in the right figure. A green checkmark indicates a token is accepted, and a cross indicates a token is rejected.}
    \label{fig:sampling-structure}
\end{figure}

\section{Methodology}
As discussed by \cite{xia2023speculative}, the shared attention mechanism used in Medusa \citep{cai2023medusa} significantly limits drafting efficiency, leading to a high discard rate of drafted tokens. Here, we introduces a drafting method that appends semantic adaptive tokens to sequences so that uses distinct attention queries for predicting each drafted token, offering a flexible approach applicable to various models without requiring structural modifications. Continuously, we designed the generation method using both greedy search and nucleus sampling, the processes are illustrated in Figure \ref{fig:sampling-structure}.

\subsection{Inference}

\subsubsection{Greedy search}

\begin{figure*}[htbp]
    \centering
    \includegraphics[width=\textwidth]{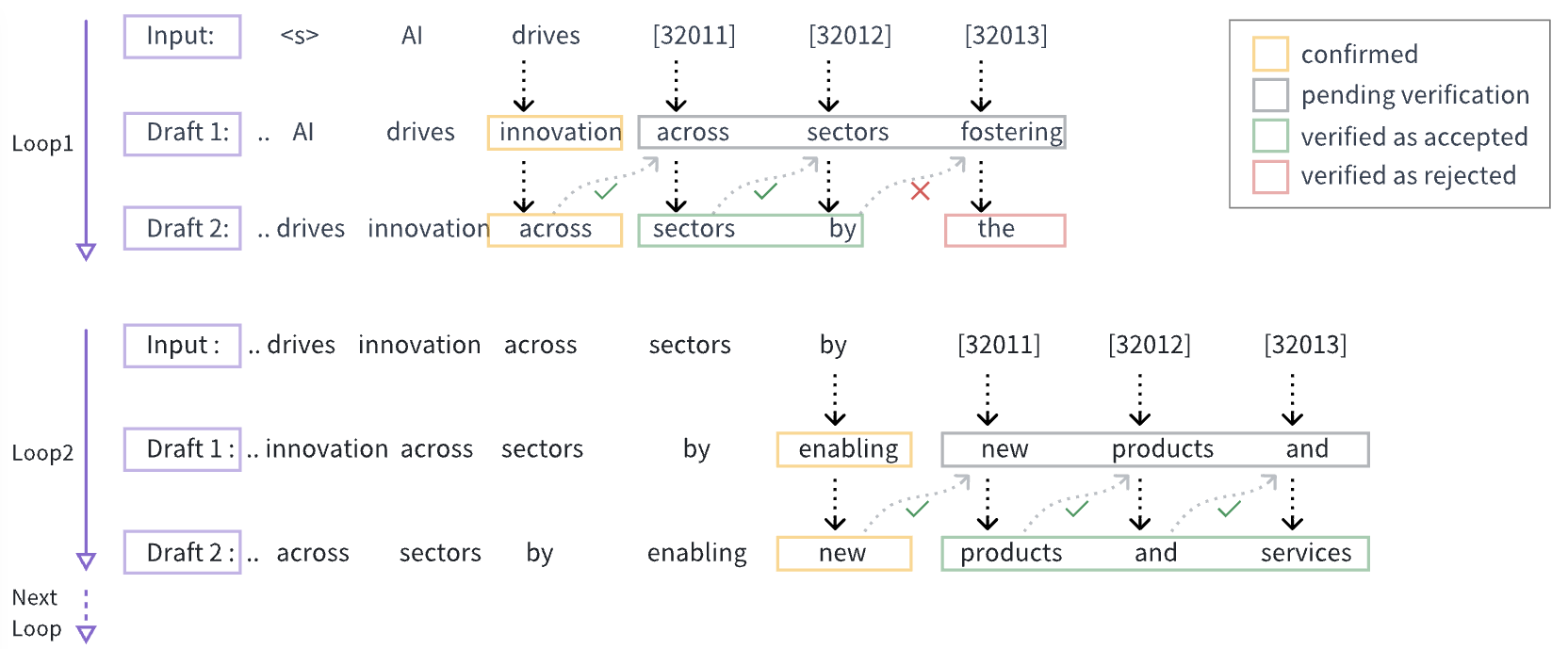}
    \caption{An example of the "two-step-draft-then-verifiy" process using the greedy search generation strategy, with [32011], [32012], [32013] as the adaptive tokens selected for the CodeLlama model. Each loop consists of three steps: after two drafting steps, the third step is verification. In the diagram, the tokens "by" and "fostering" in loop1 do not match, therefore the verified accepted token for loop1 ends at "by". The second loop passes all verifications, hence the results generated by all adaptive tokens are accepted.}
    \label{fig:greedy-demo}
\end{figure*}

For simplicity, we firstly illustrate it using a greedy search generation method as shown in left half of Figure \ref{fig:sampling-structure}. Draft step 1: Attach $k$ adaptive tokens to the sequence and generate $k$ draft tokens in parallel. Draft step 2: Leverage the outputs from step 1 to generate the second $k$ draft tokens in parallel, and then perform verification of step 3. Figure \ref{fig:greedy-demo} shows an example to illustrate the reasoning process more clearly. The inference process can be modeled as follows.

\noindent Step 1: Draft step 1
{
\begin{align}\left\{\begin{aligned}
&y_{n+1}=argmaxP_\theta(y_{n+1} | y_{1:n}, X) \\
&y_{n+2}'=argmaxP_\theta(y_{n+2}' | y_{1:n}, a_1, X) \\
&\cdots \\
&y_{n+k+1}'=argmaxP_\theta(y_{n+k+1}' | y_{1:n}, a_{1:k}, X) \\
\end{aligned}\right.\end{align}
}

\noindent Step 2: Draft step 2
{
\begin{align}\left\{\begin{aligned}
&y_{n+2}=argmaxP_\theta(y_{n+2}|y_{1:n+1}, X) \\
&y_{n+3}''=argmaxP_\theta(y_{n+3}'' | y_{1:n+1}, y_{n+2}', X) \\
&\cdots \\
&y_{n+k+2}''=argmaxP_\theta(y_{n+k+2}'' | y_{1:n+1}, y_{n+2:n+k+1}', X) \\
\end{aligned}\right.\end{align}
}

\noindent Step 3: The verification strategy
{
\begin{align}y_i=\left\{\begin{aligned}
&y_{i}, & \text{if} \ n+1 \leq i \leq n+2 \\
&y_{i}'', & \text{if} \ n+2 < i \leq n+2+k \ \text{and} \  y_{i-1}'' = y_{i-1}' \\
&\text{stop \ cur \ loop}, &  \text{otherwise} \\
\end{aligned}\right.\end{align}
}

Let $X$ represent the input of the model, which is a sequence of arbitrary length, $y$ represent the model's output and $a$ represent the semantic adaptive tokens.  Given that $y_{n+1}'$ and $y_{n+2}''$ are tokens that have been definitively accepted, they can be directly denoted as $y_{n+1}, y_{n+2}$. $y_{n+2}', y_{n+3}'',..., y_{n+k+1}', y_{n+k+2}''$ represent generated draft tokens, which need further determination on whether they can be accepted. By comparing these drafted tokens, through the Step 3 Verification process, we can obtain accepted tokens. When the stop condition of the current loop is triggered, there will be a maximum of $2+k$ accepted tokens. That is to say, if every drafted token is accepted, this will yield a maximum of $2+k$ tokens per loop with only 2 iterations, surpassing the naive implementation which would get only 2 accepted tokens.
\indent This step diverges from the process described by \cite{xia2023speculative}, where the draft phase consists of only one step before immediately moving on to the verification phase. Their method primarily relies on the higher probability outputs of the model as the criterion for accepting drafted tokens. However, it is possible for the model to confidently produce incorrect answers with high probability. Thus, relying solely on tokens probability for verification could decrease the accuracy during model inference. Our verification approach ensures that under greedy search, the decoding results are entirely equal to those from the original inference process.

\subsubsection{Nucleus sampling}
As shown in the right half of Figure \ref{fig:sampling-structure}, the nucleus sampling can also be divided into three steps.

\noindent Draft Step 1: This step is essentially similar to the method discussed previously, where several adaptive tokens are concatenated at the end of the input sequence and then sent together into the target model. The difference lies in the output for each position; due to the use of nucleus sampling, there are actually many possible draft tokens.

\noindent Draft Step 2: The draft tokens generated in the first step are organized into a tree structure, where each branch represents an alternative sequence. These alternative sequences are then presented to the model to obtain output in one go. To accurately process a given candidate sequence with a decoder only model, we introduce a manually designed tree structure, along with a corresponding attention mask within every attention layer. This approach ensures that the calculation for each token accurately reflects its relationships in draft step 1. This method, known as tree attention, was detailed in the research by \cite{he2023rest}, \cite{cai2023medusa}, and \cite{li2024eagle}.

\noindent Verification: During the verification phase, we adopt a simple judgment method similar to \cite{he2023rest}, akin to the verification strategy and stop strategy mentioned in the context of greedy search. We select the longest branch of the tree that has been verified and accepted as the result for the current loop.

\subsection{Training}

This training approach differs from the Bart-based model trained by \cite{xia2023speculative}, which employs a consistent [Mask] token for random masking in translation tasks. Given the short sequence lengths typical of translation tasks, this method can be effective. However, for longer sequences, without any optimization during the training process, the learning effectiveness of standard tokens could be adversely affected. It is preferable to adopt a new training approach, which minimizes the impact of semantic adaptive tokens on the model itself. See loss detail in Appendix \ref{appendix:training-loss}.

\noindent \textbf{Basic approach}. During training, the input sequences, originally composed entirely of standard tokens, are selectively replaced with semantic adaptive tokens, with the quantity of replacement approximated a Poisson distribution. This operation can also be considered as a mask operation and distribution of the number of adaptive tokens can be regarded as the combination of two independent random processes. Initially, 10\% of the positions in a sequence of length $n$ are selected in a uniform distribution random selection process. Subsequently, the maximum mask window size is defined as $L$, and the mask window size is determined by a uniform distribution random process, selecting a value between 1 and $L$. The combination of these two processes produces a specific probability distribution. For the entire sequence, when $n$ is large, the distribution of the number of replaced tokens approximates a Poisson distribution. Let $A$ represent the semantic adaptive tokens, this mixed sequence can be expressed as
\begin{equation}
M_{1:n} = H(Y_{1:n}, A_{1:L})
\end{equation}
After incorporating semantic adaptive tokens, for a given prediction value, training exclusively with standard tokens results in a prediction value for all inputs as 
\begin{equation}
\hat{y_i} = P_\theta(\hat{y_i}|Y_{1:i-1})
\end{equation}
By directly inserting all adaptive tokens into the sequence to obtain sequence $M$, the prediction value for the input is
\begin{equation}
\hat{y_i} = P_\theta(\hat{y_i}|M_{1:i-1})
\end{equation}
Let the loss function be denoted by $f$. For a sequence $Y$, the loss function for Auto Regression Language Models (LLMs) can be expressed as
\begin{equation}
f(\hat{Y}, Y) = -\frac{1}{n}\sum_{i=1}^nlog(\hat{y_i})    
\end{equation}
Therefore, for two different sequences, one being target sequence $Y_{M}$ and the other being the prediction value $\hat{M}$, the loss of basic approach can be expressed as
\begin{equation}
loss = f(\hat{M}, Y_{M})
\end{equation}
During the training process of LLMs, which is parallel training, the loss is generally calculated as the cross-entropy for all tokens indiscriminately. This basic approach to calculating loss significantly affects the training of standard tokens, leading to a higher loss as Figure \ref{fig:loss-label} shows.

\noindent \textbf{Improved approach}. An improved approach is to isolate the impact of adaptive tokens on standard tokens. While, during model inference, it is only necessary to append several adaptive tokens at the end of the sequence, rather than inserting them in the middle. However, for more efficient training, multiple adaptive tokens are randomly inserted in the middle of the sequence during the training process. This can affect the training of standard tokens, as the presence of adaptive tokens ahead of them may lead to increased loss for these standard tokens compared to normal circumstances.
To address this issue, two types of input sequences can be used in the actual training process. The first type consists entirely of standard tokens, leading to an output represented by $Y$. The second type mixes standard tokens with adaptive tokens, with the prediction denoted by $\hat{M}$ and corresponding labels denoted as $Y_{M}$. $M_{mask} $represents a matrix with the same shape as $M$, which corresponds to positions in the $M$ sequence that are replaced by adaptive tokens set to 1, and all other positions set to 0. The loss is then calculated as:
\begin{equation}
loss = \frac{1}{2}(f(\hat{Y}, Y) + wf(\hat{M}, Y_{M}) \odot M_{mask})
\end{equation}

In this training approach, the two types of sequences are simply set to each constitute half of the training data, with the weight $w$ set to 1.

\section{Experiments}

\subsection{Dataset}
Our training data is selected from StarCoder \citep{li2023starcoder}, chosen entirely at random in proportion to the distribution of programming languages. Additionally, we have conducted a decontamination process on training data to ensure that the test dataset will not appear in the training dataset. Following the approach used by CodeLlama \citep{roziere2023code}, we adopt an infilling objective for training our models. To elaborate, we dissect training texts into three components: a prefix, a middle section, and a suffix, by choosing split points from a uniform distribution across the document's length at the character level. We then rearrange these segments into two configurations: prefix-suffix-middle (PSM) for half of the splits and suffix-prefix-middle (SPM) for the remainder, ensuring compatibility. This reorganization is executed with a 50\% likelihood, meaning each configuration is applied with a 25\% probability.

\subsection{Training details}
In our study, we configured the finetuning parameters by referencing those outlined in the CodeLlama paper. The optimizer is AdamW \citep{Loshchilov_Hutter_2017}, with $\beta_1$ and $\beta_2$ values set at 0.9 and 0.95, respectively. Our approach includes a focused continuous pretraining phase limited to just 2000 steps, utilizing a cosine schedule with 400 warm-up steps. We set an initial learning rate of 5e-5, and a final learning rate of 1e-5. The training process leverages a batch size of 4M tokens, which are organized into sequences of 16,384 tokens each. In the subsequent experiment, we trained both CodeLlama-7B and 13B models, varying the max mask window size (L) for each to obtain multiple trained models, which we refer to as SDSAT models. The semantic adaptive tokens used here are all the same value. Models trained with diverse tokens were studied in Appendix \ref{appendix:token-impact}

\subsection{Results}
We conducted accuracy test for original CodeLlama-7B and 13B models. It is worth mentioning that when testing them on the following datasets, there are many details that can affect the accuracy, such as "stop words" and "max new tokens" of the generation, whether to do "strip operation" on prompt, etc. If these details are inconsistent, the results will also change accordingly. Since CodeLlama has not disclosed its testing specifics, re-evaluating their released models results in slight discrepancies in accuracy compared to the figures reported by them. We tentatively conclude that these minor differences due to the nuances in the testing process do not detract from the overall conclusions of our work.

\subsubsection{Python code generation}
We reference the CodeLlama paper to select our test datasets. Similarly, we begin by reporting results for Python code generation using the HumanEval \citep{chen2021evaluating} and MBPP \citep{austin2021program} benchmarks. For more precise evaluations, we use the test split of MBPP (santitized), which contains 257 problems manually verified. Additionally, we also conduct re-evaluations on the original version of CodeLlama. In this section, we only present zero-shot results. 
Table~\ref{tab:pygen} indicates that the accuracy of the 7B model is nearly identical to the native CodeLlama model. Additionally, the 13B model exhibits slightly higher overall accuracy compared to the native model.

\begin{table}
\centering
    \begin{center}
    \begin{tabular}{cccc} 
    \toprule
    \textbf{Model} & \textbf{Size} & \textbf{HumanEval}& \textbf{MBPP}\\ 
     & & \textbf{pass@1}& \textbf{pass@1}\\ 
     \midrule
     CodeLlama& 7B& 33.5\%& 49.8\%\\ 
     SDSAT (L=3) & 7B& 33.5\%& 48.6\%\\ 
     SDSAT (L=5) & 7B& 31.1\%& 49.8\%\\ 
     \midrule
     CodeLlama & 13B& 36.0\%& 51.0\%\\ 
     SDSAT (L=7) & 13B& 38.4\%& 51.4\%\\ 
     \bottomrule
    \end{tabular}
    \end{center}
    \caption{HumanEval Pass@1 scores and MBPP-sanitized Pass@1 scores with greedy search decoding.}
    \label{tab:pygen}
\end{table}

\subsubsection{Multilingual evaluation}
Next, we also evaluate our fine-tuned models on a more diverse set of programming languages using the MultiPL-E benchmark \citep{cassano2022multipl}. Just like CodeLlama, we report results for Python, C++, Java, PHP, TypeScript, C\#, and Bash in Table~\ref{tab:multigen}.

After training, the 7B model experienced a slight decrease in average accuracy, dropping from 26.3\% to 24.5\%, with the most significant decline observed in the programming language Java, C\# and Bash. In contrast, the 13B model exhibited an increase in average accuracy, particularly in the TypeScript (TS) language. It's worth noting that this could be due to inherent fluctuations in the original CodeLlama's training process for TS, where the 7B model's accuracy in TS even surpassed that of the 13B model.

In evaluating models using the HumanEval \citep{chen2021evaluating} and MultiPL-E \citep{cassano2022multipl}, we observe fluctuations in the accuracy of models saved at different stages of the training process. These fluctuations may be attributed to the limited size of the test data available for each language. Additionally, it's important to acknowledge that potential biases in our randomly selected datasets could contribute to the fluctuating performance across multiple coding languages.

\begin{table*}[htbp]
\centering
\begin{center}
\begin{tabularx}{\textwidth}{>{\raggedleft\arraybackslash}cXXXXXXXX}
\toprule
\textbf{Model} & \textbf{Size} & \textbf{C++} & \textbf{Java} & \textbf{PHP} & \textbf{TS} & \textbf{C\#} & \textbf{Bash} & \textbf{Average} \\
\midrule
CodeLlama & 7B & 28.6\% & 34.2\% & 24.2\% & 33.3\% & 25.3\% & 12.0\% & 26.3\% \\
SDSAT (L=3) & 7B & 36.7\% & 29.7\% & 25.5\% & 30.8\% & 21.5\% & 8.9\% & 25.5\% \\
SDSAT (L=5) & 7B & 32.9\% & 27.2\% & 24.2\% & 33.3\% & 20.3\% & 8.9\% & 24.5\% \\
\midrule
CodeLlama & 13B & 39.1\% & 38.0\% & 34.2\% & 29.6\% & 27.3\% & 15.2\% & 30.6\% \\
SDSAT (L=7) & 13B & 39.1\% & 37.3\% & 33.5\% & 37.1\% & 27.9\% & 13.9\% & 31.5\% \\
\bottomrule
\end{tabularx}
\end{center}
\caption{MultiPL-E Pass@1 scores with greedy search decoding across various programming languages.}
\label{tab:multigen}
\end{table*}

\subsubsection{Infilling evaluation}
\cite{allal2023santacoder} adapted the HumanEval code infilling benchmark for various programming languages through the use of MultiPL-E \citep{cassano2022multipl}. In this process, individual lines of code are masked, and the predictions are scored with an exact match metric against the ground truth solution.
From Table \ref{tab:infilling}, we can see that our fine-tuned SDSAT models and the original CodeLlama models exhibit similar performance across Python, Java, and JavaScript. Notably, in Java, the fine-tuned models demonstrate a higher accuracy. Given the large size of the test dataset, the results are relatively stable.

\begin{table}[htbp]
\centering
\begin{center}
\begin{tabularx}{\linewidth}{cXXXX}
\toprule
\textbf{Model} & \textbf{Size} & \textbf{Python} & \textbf{Java} & \textbf{JS} \\
\midrule
CodeLlama & 7B & 72.70\% & 77.60\% & 82.60\% \\
SDSAT (L=3) & 7B & 72.67\% & 78.37\% & 82.53\% \\
SDSAT (L=5) & 7B & 72.29\% & 78.99\% & 82.12\% \\
\midrule
CodeLlama & 13B & 74.50\% & 80.00\% & 85.00\% \\
SDSAT (L=7) & 13B & 73.54\% & 84.94\% & 84.36\% \\
\bottomrule
\end{tabularx}
\end{center}
\caption{Multilingual HumanEval single line infilling with MultiPL-E. Exact match rates on the line infilling benchmark from \cite{allal2023santacoder} with greedy decoding. Evaluated in suffix-prefix-middle (SPM) format only. }
\label{tab:infilling}
\end{table}

\subsection{Walltime improvement}
To test the inference performance of the 7B and 13B models, we use a portion of HumanEval \citep{chen2021evaluating} and MultiPL-E \citep{cassano2022multipl} infilling datasets as representatives for generation tasks and completion tasks respectively, and analyze their various performance metrics, summarized in Table \ref{tab:speed}. We measure walltime improvements with a batch size of 1 on a single NVIDIA A100 for both greedy search (temp=0) and nucleus sampling (temp$>$0).

\begin{table*}[htbp]
\centering
\begin{center}
\begin{tabularx}{\textwidth}{>{\raggedleft\arraybackslash}p{1cm}|p{1cm}|X|p{3cm}XXX}
\toprule
\textbf{Size} & \textbf{L} & \textbf{Temp} & \textbf{Data} & \multicolumn{2}{c}{\textbf{Tokens/s}} & \textbf{Speed} \\
 &  &  &  & \textbf{CodeLlama} & \textbf{SDSAT} &  \\
\midrule
\multirow{6}{*}{7B} & \multirow{6}{*}{5} & \multirow{2}{*}{0} & HumanEval & 37.09 & 114.92 & $>$ \textbf{3.0X} \\
 &  &  & MultiPL-E infilling & 35.93 & 110.1 & $>$ \textbf{3.0X} \\
 &  & \multirow{2}{*}{0.2} & HumanEval & 36.53 & 107.59 & $>$ 2.9X \\
 &  &  & MultiPL-E infilling & 34.24 & 87.11 & $>$ 2.5X \\
 &  & \multirow{2}{*}{1} & HumanEval & 36.66 & 81.45 & $>$ 2.2X \\
 &  &  & MultiPL-E infilling & 34.16 & 81.45 & $>$ 2.3X \\
\midrule
\multirow{6}{*}{13B} & \multirow{6}{*}{7} & \multirow{2}{*}{0} & HumanEval & 29.54 & 105.1 & $>$ \textbf{3.5X} \\
 &  &  & MultiPL-E infilling & 27.35 & 87.53 & $>$ 3.2X \\
 &  &  \multirow{2}{*}{0.2} & HumanEval & 28.63 & 99.05 & $>$ 3.4X \\
 &  &  & MultiPL-E infilling & 26.12 & 66.64 & $>$ 2.5X \\
 &  &  \multirow{2}{*}{1} & HumanEval & 28.53 & 69.84 & $>$ 2.4X \\
 &  &  & MultiPL-E infilling & 25.95 & 63.42 & $>$ 2.4X \\
\bottomrule
\end{tabularx}
\end{center}
\caption{Comparison of speed. Notably, SDSAT's selection for "Tokens per Second" (token/s) showcases the highest speed across all $k$ values. Within the nucleus sampling method, the $k$ value also corresponds to the tree-depth. It is important to note that the $k$ value is exclusive to the SDSAT model. For the CodeLlama model, the $k$ value is inherently set to 0. Additionally, a temperature setting of 0 indicates the utilization of greedy search.}
\label{tab:speed}
\end{table*}

\subsubsection{Greedy search}
\begin{figure*}[tbp]
    \centering
    \includegraphics[width=\textwidth]{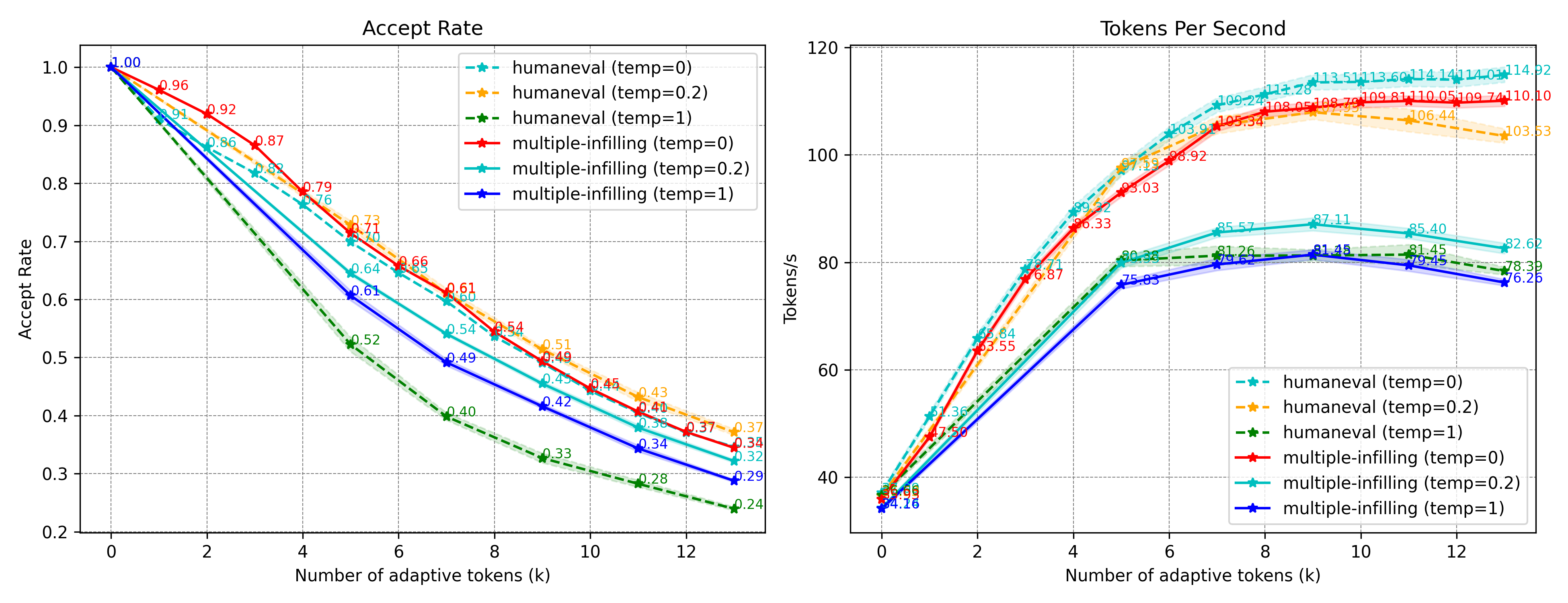}
    \caption{Performance of SDSAT-7B (L=5). Left: Accept rate, which means the average number of tokens accepted divided by the number of adaptive tokens. Right: Tokens per second of the generated new tokens.}
    \label{fig:7B-T2-inference}
\end{figure*}

In the process of inference, by controlling the use of different numbers of adaptive tokens, denoted here as $k$, we can observe the trend of changes in various performance indicators. As $k$ increases, the loop time does not increase linearly but remains almost flat, which is a significant reason why our algorithm can effectively improve inference speed. According to Figure \ref{fig:7B-T2-inference}, the "Tokens per Second" metric clearly shows that as k gradually increases, the speed continues to rise, Specifically, at $k$=13, there is an approximate 3.1X increase in speed on the HumanEval dataset \citep{chen2021evaluating}. Additionally, the "Accept Rate" aligns with our intuitive understanding that for more distant adaptive tokens, as they are further from the effective context, the probability of accurate prediction decreases.

\begin{figure*}[tbp]
    \centering
    \includegraphics[width=1\textwidth]{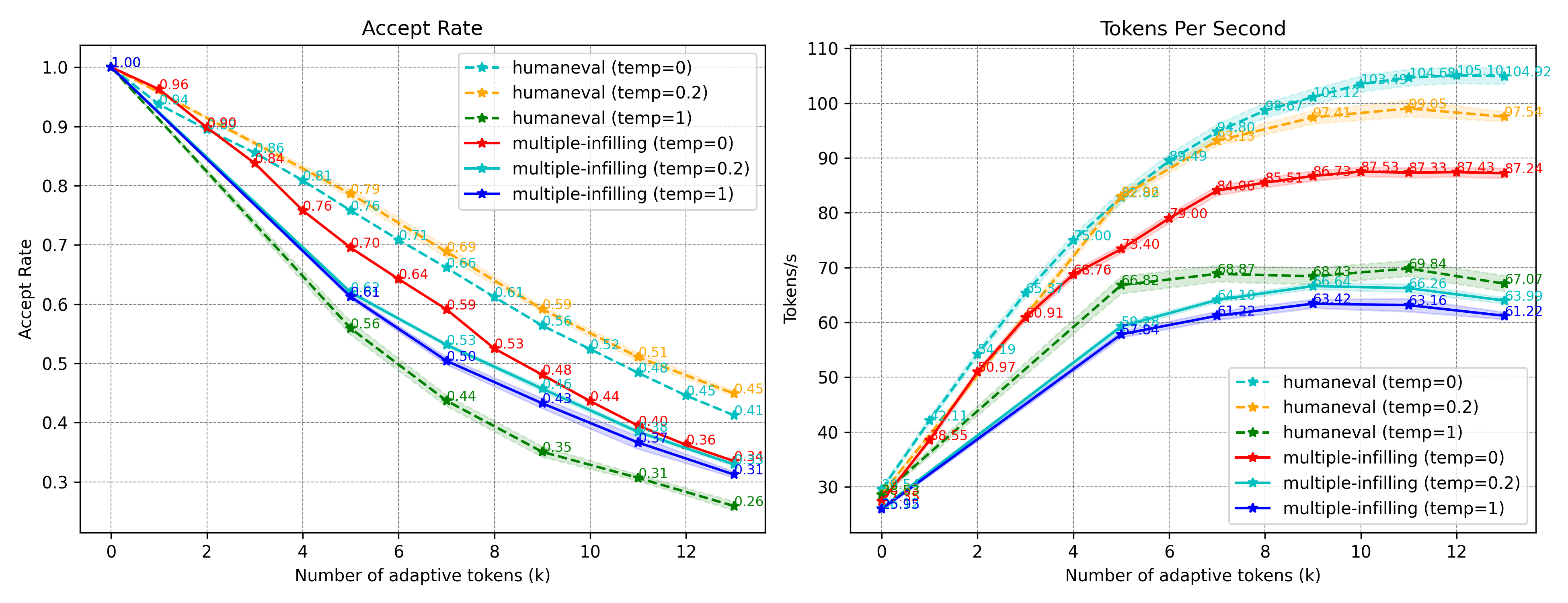}
    \caption{Performance of SDSAT-13B (L=7). Left: Accept rate. Right: Tokens per second of the generated new tokens}
    \label{fig:13B-T2-inference}
\end{figure*}

For the 13B model, as seen from Figure \ref{fig:13B-T2-inference}, due to its stronger learning capabilities, its "Accept Rate" is higher compared to the 7B model, which directly affects its speed improvement ratio. Specifically, at $k$=13, there is a more than 3.5X increase in speed on the HumanEval dataset.

\subsubsection{Nucleus sampling}

We set the parameters to top\_k=10, top\_p=0.95, and conduct experiments with two temperature settings: 0.2 and 1. For this generation approach, we design the number of adaptive tokens (tree depth) from 5 to 13 and compare it against the traditional nucleus sampling method. 

From the data presented in Figure \ref{fig:7B-T2-inference} and \ref{fig:13B-T2-inference}, it's observable that under the nucleus sampling method, the 13B model exhibits faster inference speeds compared to the 7B model. 
Additionally, a lower temperature setting results in further acceleration of inference speed. This phenomenon can be attributed to two main factors: firstly, the use of tree attention requires providing the model with longer inputs during draft step 2; more critically, employing a sampling algorithm with a higher temperature introduces greater uncertainty, which in turn lowers the acceptance rate.

Several conclusions can be drawn from the overall analysis: 1) The larger the model size, the more effective the learning outcome of semantic adaptive tokens, which is reflected in both higher acceptance rates and speed improvements. 2) As the number of adaptive tokens increases, the speed benefit becomes more pronounced. 3) Generally, a lower temperature setting results in greater speed gains.

\section{Conclusion}

This paper utilizes speculative decoding by introducing semantic adaptive tokens, enabling existing LLMs to generate accurate draft tokens with minimal cost and without any structural modifications to the model. This approach significantly enhances the generation speed of the models while maintaining nearly unchanged accuracy through optimized training. Additionally, we have designed universal greedy search and nucleus sampling schemes that can be easily transferred to other LLMs.

\bibliography{colm2024_conference}
\bibliographystyle{colm2024_conference}

\appendix

\section{Training loss}
\label{appendix:training-loss}
\begin{figure}
    \centering
    \includegraphics[width=0.7\linewidth]{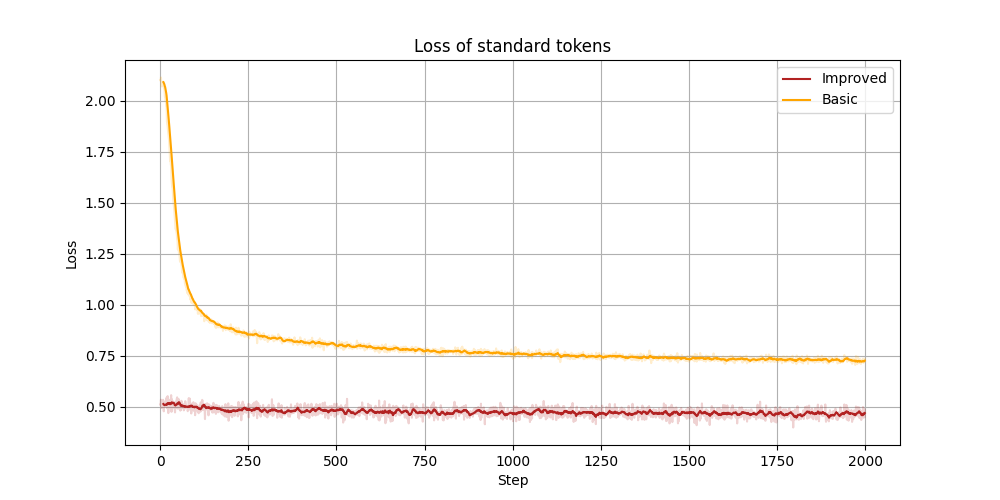}
    \caption{The loss curves of standard tokens corresponding to two different training methods. The loss of standard tokens is calculated by excluding the loss associated with adaptive tokens and computing the average loss across all standard tokens.}
    \label{fig:loss-label}
\end{figure}

Figure \ref{fig:loss-label} presents the loss curves for the two approaches. The loss of the basic training approach consistently exceeds that of the improved approach, which will affect the benchmark of the model.

\section{Adaptive token impact}
\label{appendix:token-impact}
We also conducted experiments to investigate the effects of training models with diverse semantic adaptive tokens. Our experiments indicate that using varied tokens does not significantly impact accuracy compared to using identical tokens. However, training with diverse tokens restricts us from arbitrarily extending the length of $k$; we can only use tokens with a length less than $L$. This limitation might constrain flexibility during inference. Therefore, we recommend training models with identical tokens.

\begin{figure*}[htbp]
    \centering
    \includegraphics[width=\textwidth]{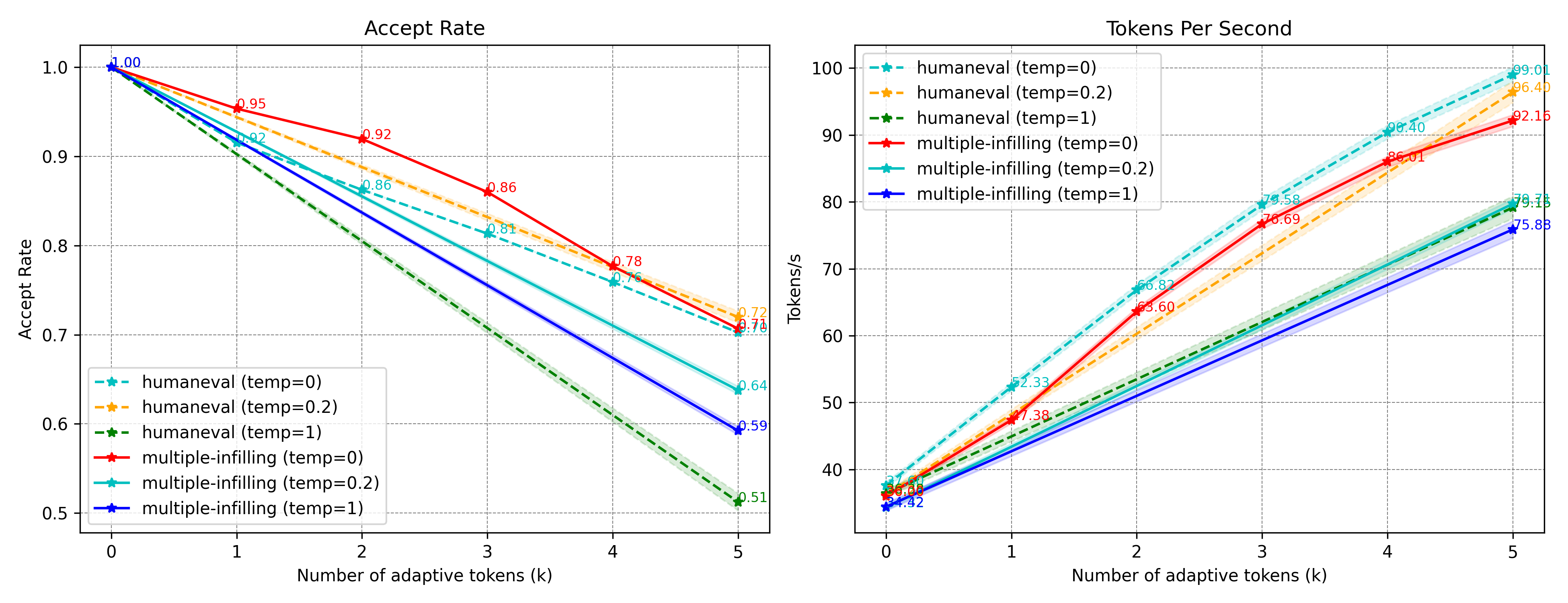}
    \caption{Performance of SDSAT-D-7B (L=5) trained with diverse adaptive tokens ([32011]-[32012] are used for CodeLlama). Left: Accept rate. Right: Tokens per second of the generated new tokens.}
    \label{fig:7B-T1-inference}
\end{figure*}

\begin{figure*}[htbp]
    \centering
    \includegraphics[width=\textwidth]{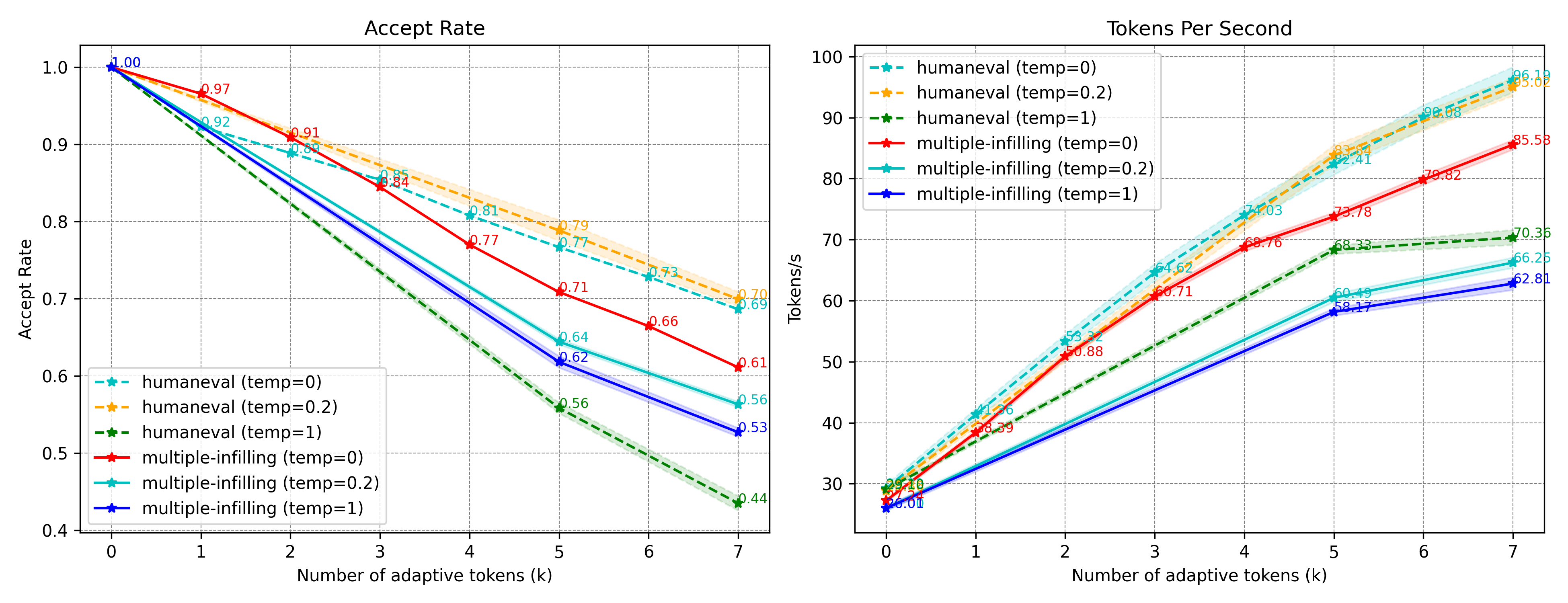}
    \caption{Performance of SDSAT-D-13B (L=7) trained with diverse adaptive tokens ([32011]-[32012], [31864] and [31857] are used for CodeLlama). Left: Accept rate. Right: Tokens per second of the generated new tokens.}
    \label{fig:13B-T1-inference}
\end{figure*}

Figures \ref{fig:7B-T1-inference} and \ref{fig:13B-T1-inference} illustrate the results obtained by employing this method, with models trained using diverse adaptive tokens represented by SDSAT-D.

\begin{table}
\centering
    \begin{center}
    \begin{tabular}{cccc} 
    \toprule
    \textbf{Model} & \textbf{Size} & \textbf{HumanEval}& \textbf{MBPP}\\ 
     & & \textbf{pass@1}& \textbf{pass@1}\\ 
     \midrule
     CodeLlama& 7B& 33.5\%& 49.8\%\\ 
     SDSAT-D (L=5) & 7B& 32.3\%& 48.3\%\\ 
     \midrule
     CodeLlama & 13B& 36.0\%& 51.0\%\\ 
     SDSAT-D (L=7) & 13B& 38.4\%& 51.0\%\\ 
     \bottomrule
    \end{tabular}
    \end{center}
    \caption{HumanEval Pass@1 scores and MBPP-sanitized Pass@1 scores.}
    \label{tab:T1-pygen}
\end{table}

\begin{table*}[htbp]
\centering
\captionsetup{justification=centering}
\begin{center}
\begin{tabularx}{\textwidth}{>{\raggedleft\arraybackslash}cXXXXXXXX}
\toprule
\textbf{Model} & \textbf{Size} & \textbf{C++} & \textbf{Java} & \textbf{PHP} & \textbf{TS} & \textbf{C\#} & \textbf{Bash} & \textbf{Average} \\
\midrule
CodeLlama & 7B & 28.6\% & 34.2\% & 24.2\% & 33.3\% & 25.3\% & 12.0\% & 26.3\% \\
SDSAT-D (L=5) & 7B & 31.7\% & 26.6\% & 27.3\% & 33.3\% & 19.6\% & 10.1\% & 24.8\% \\
\midrule
CodeLlama & 13B & 39.1\% & 38.0\% & 34.2\% & 29.6\% & 27.3\% & 15.2\% & 30.6\% \\
SDSAT-D (L=7) & 13B & 39.1\% & 36.1\% & 32.9\% & 39.0\% & 26.0\% & 14.6\% & 31.3\% \\
\bottomrule
\end{tabularx}
\end{center}
\caption{MultiPL-E Pass@1 scores. Using greedy decoding in different programming languages.}
\label{tab:T1-multigen}
\end{table*}

\begin{table}[htbp]
\centering
\begin{center}
\begin{tabularx}{\linewidth}{cXXXX}
\toprule
\textbf{Model} & \textbf{Size} & \textbf{Python} & \textbf{Java} & \textbf{JS} \\
\midrule
CodeLlama & 7B & 72.70\% & 77.60\% & 82.60\% \\
SDSAT-D (L=5) & 7B & 71.42\% & 78.31\% & 82.63\% \\
\midrule
CodeLlama & 13B & 74.50\% & 80.00\% & 85.00\% \\
SDSAT-D (L=7) & 13B & 73.53\% & 85.24\% & 84.76\% \\
\bottomrule
\end{tabularx}
\end{center}
\caption{Multilingual HumanEval single line infilling with MultiPL-E.}
\label{tab:T1-infilling}
\end{table}

\end{document}